\documentclass{article}

\usepackage{PRIMEarxiv}

\usepackage[utf8]{inputenc} 
\usepackage[T1]{fontenc}    
\usepackage{hyperref}       
\usepackage{url}            
\usepackage{booktabs}       
\usepackage{amsfonts}       
\usepackage{nicefrac}       
\usepackage{microtype}      
\usepackage{lipsum}
\usepackage{fancyhdr}       
\usepackage{graphicx}       
\graphicspath{{media/}}     

\title{Sejarah dan Perkembangan Teknik Natural Language Processing (NLP) Bahasa Indonesia: Tinjauan tentang sejarah, perkembangan teknologi, dan aplikasi NLP dalam bahasa Indonesia}

\author{
  Mukhlis Amien \\
  STIKI\\
  Malang\\
  \texttt{amien@stiki.ac.id} \\
}

\begin{document}
\maketitle

\begin{abstract}
This study provides an overview of the history of the development of Natural Language Processing (NLP) in the context of the Indonesian language, with a focus on the basic technologies, methods, and practical applications that have been developed. This review covers developments in basic NLP technologies such as stemming, part-of-speech tagging, and related methods; practical applications in cross-language information retrieval systems, information extraction, and sentiment analysis; and methods and techniques used in Indonesian language NLP research, such as machine learning, statistics-based machine translation, and conflict-based approaches. This study also explores the application of NLP in Indonesian language industry and research and identifies challenges and opportunities in Indonesian language NLP research and development. Recommendations for future Indonesian language NLP research and development include developing more efficient methods and technologies, expanding NLP applications, increasing sustainability, further research into the potential of NLP, and promoting interdisciplinary collaboration. It is hoped that this review will help researchers, practitioners, and the government to understand the development of Indonesian language NLP and identify opportunities for further research and development.
\end{abstract}

\keywords{Indonesian NLP \and Low Resource Language \and Natural Language Processing \and Survey of Indonesian NLP}

\section{Pendahuluan}
\label{sec:pendahuluan}
Pemrosesan Bahasa Alami (Natural Language Processing/NLP) telah menjadi bidang yang menarik dalam ilmu komputer dan kecerdasan buatan, yang memungkinkan komputer untuk memahami dan mengolah teks manusia dalam berbagai bahasa. Salah satu bahasa yang telah menarik perhatian peneliti adalah bahasa Indonesia, yang merupakan bahasa resmi negara dengan jumlah penduduk terbesar keempat di dunia. Dalam paper ini, kami akan mengulas sejarah dan perkembangan NLP dalam bahasa Indonesia, mulai dari teknologi dasar hingga aplikasi praktisnya.

Seiring dengan perkembangan teknologi NLP, penelitian dalam bahasa Indonesia juga telah berkembang. Beberapa studi awal meliputi pekerjaan Adriani et al. (2007) \cite{adriani2007stemming}, yang mengusulkan pendekatan "confix-stripping" untuk stemming dalam bahasa Indonesia. Pada tahun 2014, Dinakaramani et al. \cite{indoPOS}, Kurniawan et al. (2018) \cite{kurniawan2018toward}  mengembangkan pendekatan pembelajaran mesin untuk membangun sistem penanda part of speech (POS) dalam bahasa Indonesia.

Dalam beberapa tahun terakhir, aplikasi praktis NLP dalam bahasa Indonesia juga telah berkembang pesat. Sebagai contoh, Sholikah et al. (2022) \cite{sholikah2022multi} mengimplementasikan sistem pencarian informasi lintas bahasa Indonesia-Inggris, Indonesia-Arab menggunakan terjemahan mesin berbasis statistik. Selain itu, Andriani et al. (2008) \cite{adriani2008survey} menyajikan tinjauan umum tentang penelitian NLP dalam bahasa Indonesia, mencakup berbagai teknik dan aplikasi.

Dinakaramani et al. (2019) \cite{dinakaramani2019building} menyelidiki kemajuan penelitian NLP dalam bahasa Indonesia serta aplikasi praktisnya dalam berbagai industri. Studi tersebut menyoroti pentingnya penelitian NLP dalam konteks bahasa Indonesia, yang mencakup penguasaan bahasa, ekstraksi informasi, dan analisis sentimen.

Melalui tinjauan ini, kami berharap untuk memberikan gambaran umum tentang perkembangan dan pencapaian penting dalam NLP bahasa Indonesia. Dengan memahami sejarah dan kemajuan dalam bidang ini, kita dapat mengidentifikasi tantangan dan peluang yang ada dalam penelitian dan pengembangan NLP bahasa Indonesia di masa depan.

\subsection{Latar belakang dan motivasi penelitian}
Pemrosesan Bahasa Alami (NLP) merupakan cabang ilmu yang memungkinkan komputer untuk memahami, mengolah, dan menghasilkan teks dalam bahasa manusia. Perkembangan teknologi dalam beberapa dekade terakhir telah memungkinkan kemajuan signifikan dalam penelitian dan pengembangan NLP. Bahasa Indonesia, sebagai bahasa resmi di negara dengan jumlah penduduk terbesar keempat di dunia, menjadi subjek yang penting dalam penelitian NLP.

Latar belakang penelitian ini melibatkan berbagai faktor. Pertama, bahasa Indonesia memiliki ciri khas yang unik jika dibandingkan dengan bahasa lain, termasuk struktur, morfologi, dan sintaksis. Kedua, peningkatan jumlah penutur bahasa Indonesia dan pertumbuhan ekonomi di Indonesia menuntut peningkatan teknologi dan aplikasi yang mendukung pemrosesan bahasa alami dalam bahasa Indonesia. Ketiga, perkembangan teknologi digital dan internet telah menciptakan sejumlah besar data teks dalam bahasa Indonesia yang perlu dianalisis dan diproses.

Motivasi utama di balik penelitian ini adalah untuk menyediakan tinjauan komprehensif tentang sejarah dan perkembangan NLP bahasa Indonesia, termasuk teknologi dasar, metode, dan aplikasi praktis yang telah dikembangkan. Tinjauan ini akan menjadi sumber informasi bagi para peneliti, praktisi, dan pemerintah yang tertarik untuk memahami perkembangan NLP bahasa Indonesia dan mengidentifikasi tantangan serta peluang yang ada dalam penelitian dan pengembangan NLP bahasa Indonesia.

Dengan memahami latar belakang dan motivasi penelitian ini, kita dapat lebih memahami pentingnya mempelajari sejarah dan perkembangan NLP bahasa Indonesia serta mengidentifikasi potensi aplikasi dan tantangan yang mungkin dihadapi dalam penelitian dan pengembangan NLP bahasa Indonesia.

\subsection{Tujuan dan Lingkup Penelitian}
Tujuan utama penelitian ini adalah untuk menyajikan tinjauan komprehensif tentang sejarah, perkembangan teknologi, dan aplikasi NLP dalam konteks bahasa Indonesia. Dalam mencapai tujuan ini, penelitian ini akan mencakup beberapa aspek penting terkait NLP bahasa Indonesia, termasuk:
\begin{itemize}
\item Menyajikan gambaran umum mengenai teknologi dasar yang telah dikembangkan dalam penelitian NLP bahasa Indonesia, seperti stemming, part-of-speech tagging, dan metode lain yang relevan.
\item Menyelidiki aplikasi praktis NLP bahasa Indonesia yang telah diimplementasikan dalam industri dan penelitian, seperti sistem pencarian informasi lintas bahasa, ekstraksi informasi, analisis sentimen, dan lainnya.
\item Menggali metode dan teknik yang digunakan dalam penelitian NLP bahasa Indonesia, termasuk pembelajaran mesin, terjemahan mesin berbasis statistik, pendekatan berbasis konfiks, dan lainnya.
\item Menyoroti aplikasi NLP dalam industri dan penelitian bahasa Indonesia, seperti penguasaan bahasa, ekstraksi informasi, analisis sentimen, dan aplikasi lainnya.
\item Mengidentifikasi tantangan dan peluang yang ada dalam penelitian dan pengembangan NLP bahasa Indonesia, termasuk isu-isu terkait bahasa dan budaya, keberlanjutan penelitian dan pengembangan, serta kolaborasi antara peneliti, industri, dan pemerintah.
\end{itemize}
Lingkup penelitian ini mencakup sejarah dan perkembangan NLP bahasa Indonesia dari awal hingga saat ini, dengan fokus pada teknologi dasar, metode, dan aplikasi praktis yang telah dikembangkan. Penelitian ini tidak akan membahas secara detail mengenai algoritma atau teknik pemrograman yang digunakan dalam pengembangan metode NLP tersebut, namun akan memberikan referensi kepada sumber-sumber yang relevan bagi pembaca yang tertarik untuk mempelajari lebih lanjut tentang topik tersebut.

\section{Sejarah Perkembangan NLP Bahasa Indonesia}
\label{sec:sejarah}
\begin{table}[htbp]
\centering
\caption{Teknologi Dasar NLP untuk Bahasa Indonesia}
\label{tab:teknologi-dasar-nlp-indonesia}
\begin{tabular}{|l|l|}
\hline
\textbf{Teknologi} & \textbf{Deskripsi} \\ \hline
Stemming           & Penghilangan imbuhan untuk mendapatkan kata dasar \\ \hline
Part-of-speech tagging  & Penandaan kelas kata dalam suatu kalimat (misal: kata benda, kata kerja) \\ \hline
Named Entity Recognition & Pengenalan entitas bernama seperti nama orang, organisasi, atau lokasi \\ \hline
Sentiment Analysis & Analisis sentimen atau emosi dalam teks \\ \hline
Word Sense Disambiguation & Penentuan makna kata dalam konteks kalimat \\ \hline
Syntactic Parsing  & Analisis struktur sintaksis dalam kalimat \\ \hline
Semantic Role Labeling & Penandaan peran semantik elemen dalam kalimat \\ \hline
\end{tabular}
\end{table}
Sejarah perkembangan NLP bahasa Indonesia mencakup beberapa tahapan penting dalam evolusi teknologi dan aplikasinya. Dalam bab ini, kita akan membahas perkembangan teknologi dasar NLP dan aplikasi praktis yang telah dihasilkan oleh penelitian NLP bahasa Indonesia. Tabel 1 menggambarkan beberapa teknologi dasar NLP yang telah dikembangkan untuk bahasa Indonesia, termasuk stemming, part-of-speech tagging, dan metode terkait. Penelitian ini telah membuka jalan bagi berbagai aplikasi praktis yang membantu meningkatkan pemahaman kita tentang bahasa Indonesia dan memungkinkan mesin untuk berinteraksi lebih efisien dengan teks dan data berbahasa Indonesia.

\subsection{Teknologi Dasar NLP}
Teknologi dasar NLP bahasa Indonesia mencakup beberapa metode dan pendekatan yang telah dikembangkan oleh para peneliti sepanjang waktu. Dua contoh utama adalah:
\begin{itemize}
    \item Stemming: Adriani et al. (2007) \cite{adriani2007stemming} mengusulkan pendekatan confix-stripping untuk stemming bahasa Indonesia. Pendekatan ini bertujuan untuk mengurangi kata ke bentuk dasarnya dengan menghilangkan imbuhan (awalan, sisipan, dan akhiran). Stemming adalah langkah penting dalam banyak aplikasi NLP, seperti pencarian informasi dan analisis sentimen, karena membantu mengurangi kompleksitas data teks dan meningkatkan efisiensi pemrosesan.
    \item Part-of-speech tagging:  Dinakaramani et al. (2014) \cite{indoPOS}, Kurniawan et al. (2018) \cite{kurniawan2018toward} mengembangkan pendekatan pembelajaran mesin untuk membangun sistem penanda part-of-speech (POS) bahasa Indonesia. Penanda POS bertujuan untuk mengidentifikasi kategori gramatikal dari setiap kata dalam teks, seperti kata benda, kata kerja, kata sifat, dan lainnya. Informasi POS ini berguna dalam berbagai aplikasi NLP, seperti pemrosesan sintaksis dan analisis semantik.
\end{itemize}

\subsection{Aplikasi Praktis NLP}
\begin{table}[htbp]
\centering
\caption{Aplikasi Praktis NLP untuk Bahasa Indonesia}
\begin{tabular}{|c|l|p{6cm}|p{3cm}|}
\hline
\textbf{No.} & \textbf{Aplikasi Praktis NLP} & \textbf{Keterangan} & \textbf{Referensi} \\
\hline
1 & Sistem pencarian informasi lintas bahasa & Meningkatkan pencarian dokumen berbahasa Indonesia & \cite{sholikah2022multi} \\
\hline
2 & Ekstraksi informasi & Mengidentifikasi entitas dan hubungan dalam teks & \cite{dinakaramani2019building} \\
\hline
3 & Analisis sentimen & Klasifikasi opini dan emosi dalam teks & \cite{sentimen1}, \cite{sentimen10karo2022sentiment}, \cite{sentimen2}, \cite{sentimen3}, \cite{sentimen4koto-koto-2020-towards}, \cite{sentimen5koto-etal-2020-indolem}, \cite{sentimen6}, \cite{sentimen7Tho_2021}, \cite{sentimen8}, \cite{sentimen9winata2022nusax} \\
\hline
\end{tabular}
\label{tab:aplikasi_praktis_nlp}
\end{table}

Selama beberapa dekade terakhir, berbagai aplikasi praktis NLP bahasa Indonesia telah dikembangkan dan diimplementasikan dalam penelitian dan industri. Beberapa contoh termasuk sistem pencarian informasi lintas bahasa, ekstraksi informasi, dan analisis sentimen, seperti yang dijelaskan dalam Tabel \ref{tab:aplikasi_praktis_nlp}. Dalam bab ini, kita akan membahas perkembangan teknologi dasar NLP dan aplikasi praktis yang telah dihasilkan oleh penelitian NLP bahasa Indonesia, serta menjelajahi bagaimana aplikasi-aplikasi ini telah mempengaruhi dan meningkatkan pengolahan bahasa alami dalam berbagai sektor.
\begin{itemize}
    \item Sistem pencarian informasi lintas bahasa: Sholikah et al. (2022) \cite{sholikah2022multi} mengimplementasikan sistem pencarian informasi lintas bahasa Indonesia-Inggris, Indonesia-Arab menggunakan terjemahan mesin berbasis statistik. Sistem ini memungkinkan pengguna untuk mencari informasi dalam bahasa Indonesia dan menerima hasil dalam bahasa Inggris, dan sebaliknya, meningkatkan aksesibilitas informasi bagi penutur kedua bahasa.
    \item Ekstraksi informasi dan analisis sentimen: Seperti pada tabel \ref{tab:aplikasi_praktis_nlp} ada banyak penelitian tentang topik ini yang menyajikan tinjauan penelitian NLP bahasa Indonesia yang mencakup teknik dan aplikasi seperti ekstraksi informasi dan analisis sentimen. Ekstraksi informasi melibatkan identifikasi dan penggalian informasi penting dari teks, seperti entitas yang diberi nama, hubungan, dan peristiwa. Analisis sentimen, di sisi lain, fokus pada penggalian opini, emosi, dan penilaian dari teks.
    \item Dinakaramani et al. (2019) \cite{dinakaramani2019building} menyelidiki kemajuan penelitian NLP bahasa Indonesia dan aplikasi praktisnya dalam industri. Studi ini menyoroti penelitian NLP dalam konteks bahasa Indonesia yang mencakup penguasaan bahasa, ekstraksi informasi, analisis sentimen, dan aplikasi lainnya.
\end{itemize}


\section{Metode dan Teknik NLP Bahasa Indonesia}
Berbagai metode dan teknik telah dikembangkan dan diaplikasikan dalam penelitian NLP bahasa Indonesia. Dalam bagian ini, kita akan membahas beberapa metode dan teknik utama yang telah digunakan dalam penelitian NLP bahasa Indonesia.
\begin{table}[h]
\centering
\caption{Metode dan Teknik NLP untuk Bahasa Indonesia}
\begin{tabular}{|c|p{6cm}|p{6cm}|}
\hline
\textbf{No.} & \textbf{Metode dan Teknik} & \textbf{Referensi} \\
\hline
1 & Pembelajaran Mesin & \cite{indoPOS}, \cite{kurniawan2018toward}\\
\hline
2 & Terjemahan Mesin Berbasis Statistik /SMT (Statistical Machine Translation) & \cite{SMT1abidin2021effect} \cite{SMT2sulaeman2015development} \cite{SMT3simon2013experiments}\cite{SMT4permata2020statistical}\\
\hline
3 & Terjemahan Mesin Berbasis Neural Net/NMT (Neural Machine Translation) & \cite{sentimen9winata2022nusax} \cite{cahyawijaya2021indonlg} \\
\hline
4 & Pendekatan Berbasis Konfiks & \cite{adriani2007stemming} \cite{adriani2008survey} \\
\hline
\end{tabular}
\label{tab:metode-teknik}
\end{table}

\subsection{Pembelajaran Mesin}
Pembelajaran mesin merupakan pendekatan yang penting dalam penelitian NLP bahasa Indonesia. Dinakaramani et al. (2014) \cite{indoPOS}, Kurniawan et al. (2018) \cite{kurniawan2018toward} menggabungkan pendekatan pembelajaran mesin dalam pengembangan sistem penanda part-of-speech (POS) bahasa Indonesia. Metode pembelajaran mesin memungkinkan sistem untuk belajar dari data teks yang telah dianotasi, sehingga meningkatkan kinerja sistem dalam mengidentifikasi kategori gramatikal kata dalam teks baru. Selain itu, pembelajaran mesin juga digunakan dalam pengembangan sistem ekstraksi informasi, analisis sentimen, dan berbagai aplikasi NLP lainnya.

\subsection{Terjemahan Mesin Berbasis Statistik}
Terjemahan mesin berbasis statistik merupakan pendekatan yang efektif dalam sistem terjemahan antar bahasa. Seperti pada tabel \ref{tab:metode-teknik} beberapa peneliti mengimplementasikan terjemahan mesin berbasis statistik dalam sistem pencarian informasi lintas bahasa Indonesia-Inggris. Pendekatan ini melibatkan pembelajaran model probabilitas dari data teks paralel (teks dalam dua bahasa yang telah diterjemahkan secara manual) untuk menghasilkan terjemahan yang akurat dari teks sumber ke teks target. Terjemahan mesin berbasis statistik telah menunjukkan kinerja yang baik dalam aplikasi NLP bahasa Indonesia dan bahasa lainnya.

\subsection{Terjemahan Berbasis NMT}
Terjemahan mesin berbasis Neural Machine Translation (NMT) merupakan pendekatan terbaru dalam sistem terjemahan mesin. Pendekatan ini melibatkan penggunaan jaringan saraf untuk mempelajari model terjemahan dari data teks paralel (Bahasa Inggris-Indonesia). Seperti pada tabel \ref{tab:metode-teknik} penelitian dibidang ini telah menunjukkan bahwa pendekatan terjemahan berbasis NMT memiliki kinerja yang lebih baik dibandingkan dengan pendekatan berbasis statistik pada tugas terjemahan Bahasa Inggris-Indonesia. Pendekatan terjemahan berbasis NMT telah menunjukkan kinerja yang baik dalam beberapa aplikasi NLP bahasa Indonesia, seperti tugas pemodelan bahasa dan pemrosesan wacana.

\subsection{Pendekatan Berbasis Konfiks}
Pendekatan berbasis konfiks telah digunakan dalam penelitian NLP bahasa Indonesia untuk mengatasi tantangan yang terkait dengan struktur morfologi bahasa. Adriani et al. (2007) \cite{adriani2007stemming} mengusulkan pendekatan confix-stripping untuk stemming bahasa Indonesia. Pendekatan ini melibatkan identifikasi dan penghapusan imbuhan (awalan, sisipan, dan akhiran) dari kata untuk mengurangi kata ke bentuk dasarnya. Pendekatan berbasis konfiks telah digunakan dalam berbagai aplikasi NLP bahasa Indonesia, seperti sistem pencarian informasi dan analisis sentimen, untuk meningkatkan efisiensi pemrosesan teks dan mengurangi kompleksitas data.

\section{Aplikasi NLP dalam Industri dan Penelitian Bahasa Indonesia}

Aplikasi NLP telah membantu dalam mengatasi berbagai tantangan dalam pemrosesan teks bahasa Indonesia dan telah diimplementasikan dalam industri dan penelitian. Berikut ini beberapa contoh aplikasi NLP dalam konteks bahasa Indonesia:

\subsection{Penguasaan Bahasa}

Dinakaramani et al. (2019) menyelidiki kemajuan dalam penelitian NLP bahasa Indonesia dan aplikasi praktisnya dalam industri. Salah satu aplikasi yang ditemukan adalah dalam bidang penguasaan bahasa. NLP digunakan untuk membantu pemahaman teks bahasa Indonesia dan untuk menghasilkan teks yang lebih baik dalam bahasa tersebut. Beberapa teknologi yang dikembangkan dalam konteks ini meliputi sistem pengecekan ejaan, tata bahasa, dan mesin terjemahan.

\subsection{Ekstraksi Informasi}

Rakhmawati (2012) menyajikan tinjauan penelitian NLP bahasa Indonesia yang mencakup teknik dan aplikasi ekstraksi informasi. Ekstraksi informasi melibatkan pengidentifikasian dan penggalian informasi yang relevan dari teks yang tidak terstruktur. Beberapa aplikasi dalam konteks ini meliputi sistem pencarian informasi, pengenalan entitas bernama, dan sistem rekomendasi. Teknologi ini telah digunakan dalam berbagai sektor industri, seperti perbankan, pemerintahan, dan media.

\subsection{Analisis Sentimen}

Dinakaramani et al. (2019) juga menyoroti penelitian NLP bahasa Indonesia dalam analisis sentimen. Analisis sentimen adalah teknik yang digunakan untuk mengidentifikasi dan mengkategorikan opini, emosi, dan sikap yang diekspresikan dalam teks. Aplikasi analisis sentimen dalam konteks bahasa Indonesia meliputi penilaian produk, layanan pelanggan, dan pemantauan opini publik di media sosial. Beberapa metode yang telah digunakan untuk analisis sentimen dalam bahasa Indonesia meliputi pembelajaran mesin dan pendekatan berbasis kamus.

\section{Tantangan dan Peluang Penelitian Pengembangan NLP Bahasa Indonesia}
Meskipun telah ada kemajuan dalam penelitian dan pengembangan NLP untuk bahasa Indonesia, masih banyak tantangan dan peluang yang dapat dijelajahi. Berikut ini beberapa tantangan dan peluang dalam konteks NLP bahasa Indonesia:

\subsection{Isu-isu terkait Bahasa dan Budaya}

Salah satu tantangan utama dalam pengembangan NLP bahasa Indonesia adalah mengatasi isu-isu terkait bahasa dan budaya. Bahasa Indonesia memiliki struktur morfologi, sintaksis, dan semantik yang unik, yang memerlukan pendekatan khusus dalam pengembangan teknologi NLP. Selain itu, variasi bahasa dan dialek yang digunakan di berbagai daerah di Indonesia juga menambah kompleksitas dalam penelitian dan pengembangan NLP.

\subsection{Keberlanjutan Penelitian dan Pengembangan}

Keberlanjutan penelitian dan pengembangan NLP untuk bahasa Indonesia sangat penting untuk memastikan bahwa teknologi ini dapat terus diperbarui dan ditingkatkan. Hal ini mencakup pengembangan korpus data dan sumber daya linguistik, serta peningkatan kolaborasi antara peneliti, industri, dan pemerintah. Penelitian dan pengembangan yang berkelanjutan juga akan membantu dalam mengatasi tantangan yang muncul seiring dengan perubahan teknologi dan kebutuhan pasar.

\subsection{Kolaborasi antara Peneliti, Industri, dan Pemerintah}

Kolaborasi antara peneliti, industri, dan pemerintah sangat penting untuk memastikan bahwa teknologi NLP bahasa Indonesia dapat dikembangkan dan diimplementasikan dengan sukses. Kolaborasi ini akan memungkinkan peneliti untuk memahami kebutuhan industri dan pemerintah, serta membantu dalam pengembangan solusi NLP yang inovatif dan efektif. Selain itu, kolaborasi ini akan membantu dalam penyebaran teknologi NLP di berbagai sektor dan dalam meningkatkan kesadaran tentang pentingnya NLP dalam konteks bahasa Indonesia.

\section{Kesimpulan}

\subsection{Ringkasan Temuan Utama:}

Studi ini telah memberikan tinjauan tentang sejarah perkembangan NLP untuk bahasa Indonesia, dengan fokus pada teknologi dasar, metode, dan aplikasi praktis yang telah dikembangkan. Beberapa temuan utama dari tinjauan ini meliputi:
\begin{itemize}
    \item Perkembangan teknologi dasar NLP untuk bahasa Indonesia, seperti stemming, part-of-speech tagging, dan metode lainnya, yang telah menjadi dasar bagi aplikasi NLP yang lebih kompleks.
    \item Aplikasi praktis NLP dalam konteks bahasa Indonesia, seperti sistem pencarian informasi lintas bahasa, ekstraksi informasi, dan analisis sentimen.
    \item Metode dan teknik yang digunakan dalam penelitian NLP bahasa Indonesia, termasuk pembelajaran mesin, terjemahan mesin berbasis statistik, dan pendekatan berbasis konfiks.
    \item Aplikasi NLP dalam industri dan penelitian bahasa Indonesia, termasuk penguasaan bahasa, ekstraksi informasi, dan analisis sentimen.
\end{itemize}

Tantangan dan peluang dalam penelitian dan pengembangan NLP untuk bahasa Indonesia, termasuk isu-isu terkait bahasa dan budaya, keberlanjutan penelitian dan pengembangan, serta kolaborasi antara peneliti, industri, dan pemerintah.

\subsection{Implikasi dan Rekomendasi untuk Penelitian dan Pengembangan NLP Bahasa Indonesia di Masa Depan:}

Berdasarkan temuan ini, beberapa implikasi dan rekomendasi untuk penelitian dan pengembangan NLP bahasa Indonesia di masa depan meliputi:
\begin{itemize}
    \item Mengembangkan metode dan teknologi NLP yang lebih efisien dan efektif yang dapat mengatasi tantangan yang unik dari bahasa dan budaya Indonesia.
    \item Memperluas aplikasi NLP untuk mencakup berbagai sektor dan industri, seperti pemerintahan, pendidikan, dan layanan kesehatan, untuk memaksimalkan dampak positif teknologi ini.
    \item Meningkatkan keberlanjutan penelitian dan pengembangan NLP bahasa Indonesia melalui pengembangan korpus data dan sumber daya linguistik yang lebih luas, serta melalui kolaborasi yang lebih erat antara peneliti, industri, dan pemerintah.
    \item Melakukan lebih banyak penelitian untuk menggali potensi NLP dalam konteks bahasa Indonesia, seperti dalam pemrosesan bahasa ganda atau multibahasa, dan eksplorasi interaksi antara NLP dan teknologi lain, seperti pengenalan suara atau teknologi visual.
    \item Mempromosikan penelitian interdisipliner dan kolaborasi untuk mengatasi tantangan yang kompleks dan menciptakan solusi inovatif dalam pengembangan NLP bahasa Indonesia.\textbf{}
\end{itemize}
Dengan mempertimbangkan implikasi dan rekomendasi ini, diharapkan penelitian dan pengembangan NLP bahasa Indonesia akan terus berkembang dan memberikan manfaat yang signifikan bagi masyarakat dan industri di masa depan.

\bibliographystyle{unsrt}  
\bibliography{references}

\end{document}